\pdfoutput=1
\documentclass[11pt,a4paper]{article}
\usepackage[hyperref]{acl2018}
\usepackage{times}
\usepackage{latexsym}
\usepackage{epsfig}
\usepackage{graphicx}
\usepackage{amsmath}
\usepackage{amssymb}
\newcommand{\ti}{\textit}
\usepackage{multirow}
\usepackage{url}

\aclfinalcopy % Uncomment this line for the final submission
\title{On the Automatic Generation of Medical Imaging Reports}

\author{Baoyu Jing\textsuperscript{$\dag$*} \quad  Pengtao Xie\textsuperscript{$\dag$*} \quad Eric P. Xing\textsuperscript{$\dag$}\\
  \textsuperscript{$\dag$}Petuum Inc, USA \\
\textsuperscript{*}School of Computer Science, Carnegie Mellon University, USA\\
  {\tt \{baoyu.jing, pengtao.xie, eric.xing\}@petuum.com} \\}

\date{}

\begin{document}

\maketitle

%%%%%%%%% ABSTRACT
\begin{abstract}
Medical imaging is widely used in clinical practice for diagnosis and treatment. Report-writing can be error-prone for unexperienced physicians, and time-consuming and tedious for experienced physicians. To address these issues, we study the automatic generation of medical imaging reports. This task presents several challenges. First, a complete report contains multiple heterogeneous forms of information, including \emph{findings} and \emph{tags}. Second, abnormal regions in medical images are difficult to identify. Third, the reports are typically long, containing multiple sentences. To cope with these challenges, we (1) build a multi-task learning framework which jointly performs the prediction of tags and the generation of paragraphs, (2) propose a co-attention mechanism to localize regions containing abnormalities and generate narrations for them, (3) develop a hierarchical LSTM model to generate long paragraphs. We demonstrate the effectiveness of the proposed methods on two publicly available datasets.
\end{abstract}

%%%%%%%%% BODY TEXT
\section{Introduction}
\begin{figure}[t]
  \centering
  \footnotesize
  \includegraphics[width=0.5\textwidth]{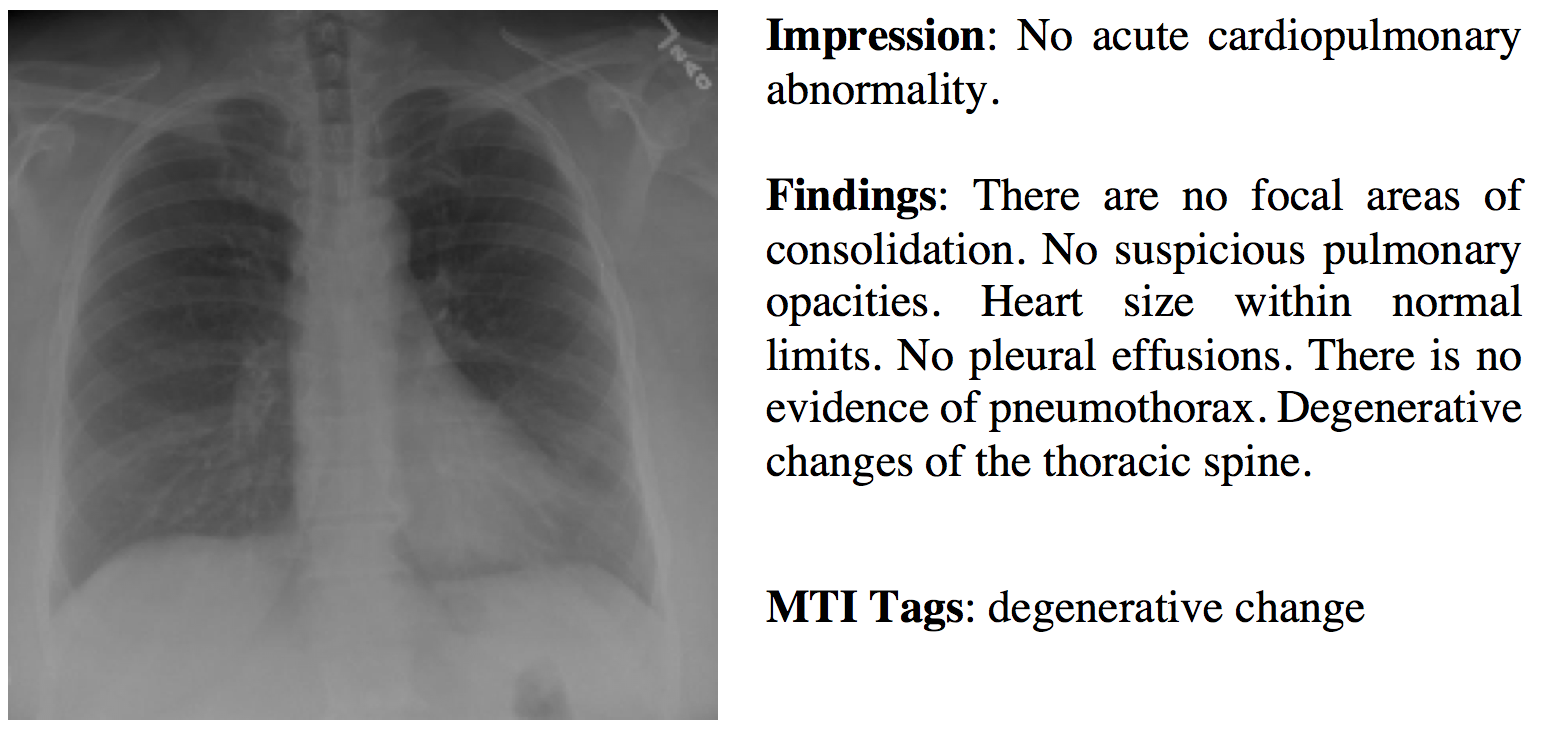}
  \caption{An exemplar chest x-ray report. In the \emph{impression} section, the radiologist provides a diagnosis. The \emph{findings} section lists the radiology observations regarding each area of the body examined in the imaging study. The \emph{tags} section lists the keywords which represent the critical information in the findings. These keywords are identified using the Medical Text Indexer (MTI).
}
\label{fig:example}
% \vspace{-0.5cm}
\end{figure}

Medical images, such as radiology and pathology images, are widely used in hospitals for the diagnosis and treatment of many diseases, such as pneumonia and pneumothorax. The reading and interpretation of medical images are usually conducted by specialized medical professionals. For example, radiology images are read by radiologists. They write textual reports (Figure \ref{fig:example}) to narrate the findings regarding each area of the body examined in the imaging study, specifically whether each area was found to be normal, abnormal or potentially abnormal.  

For less-experienced radiologists and pathologists, especially those working in the rural area where the quality of healthcare is relatively low, writing medical-imaging reports is demanding. For instance, to correctly read a chest x-ray image, the following skills are needed \cite{delrue2011difficulties}: (1) thorough knowledge of the normal anatomy of the thorax, and the basic physiology of chest diseases; (2) skills of analyzing the radiograph through a fixed pattern; (3) ability of evaluating the evolution over time; (4) knowledge of clinical presentation and history; (5) knowledge of the correlation with other diagnostic results (laboratory results, electrocardiogram, and respiratory function tests).

For experienced radiologists and pathologists, writing imaging reports is tedious and time-consuming. In nations with large population such as China, a radiologist may need to read hundreds of radiology images per day. Typing the findings of each image into computer takes about 5-10 minutes, which occupies most of their working time. In sum, for both unexperienced and experienced medical professionals, writing imaging reports is unpleasant. 

This motivates us to investigate whether it is possible to automatically generate medical image reports. Several challenges need to be addressed. First, a complete diagnostic report is comprised of multiple heterogeneous forms of information. As shown in Figure~\ref{fig:example}, the report for a chest x-ray contains \emph{impression} which is a sentence, \emph{findings} which are a paragraph, and \emph{tags} which are a list of keywords. Generating this heterogeneous information in a unified framework is technically demanding. We address this problem by building a multi-task framework, which treats the prediction of tags as a multi-label classification task, and treats the generation of long descriptions as a text generation task. 

Second, how to localize image-regions and attach the right description to them are challenging. We solve these problems by introducing a co-attention mechanism, which simultaneously attends to images and predicted tags and explores the synergistic effects of visual and semantic information. 

Third, the descriptions in imaging reports are usually long, containing multiple sentences. Generating such long text is highly nontrivial. Rather than adopting a single-layer LSTM~\cite{hochreiter1997long}, which is less capable of modeling long word sequences, we leverage the compositional nature of the report and adopt a hierarchical LSTM to produce long texts. Combined with the co-attention mechanism, the hierarchical LSTM first generates high-level topics, and then produces fine-grained descriptions according to the topics.

Overall, the main contributions of our work are:
\begin{itemize}
\item We propose a multi-task learning framework which can simultaneously predict the tags and generate the text descriptions.
\item We introduce a co-attention mechanism for localizing sub-regions in the image and generating the corresponding descriptions.
\item We build a hierarchical LSTM to generate long paragraphs.
\item We perform extensive experiments to show the effectiveness of the proposed methods.
\end{itemize}

 The rest of the paper is organized as follows. Section \ref{related_work} reviews related works. Section \ref{methods} introduces the method. Section \ref{experiments} present the experimental results and Section \ref{conclusion} concludes the paper. 

%!TEX root = ./acl_2018.tex
\pdfoutput=1
\begin{figure*}[t]
  \centering
  \footnotesize
  \includegraphics[width=\textwidth]{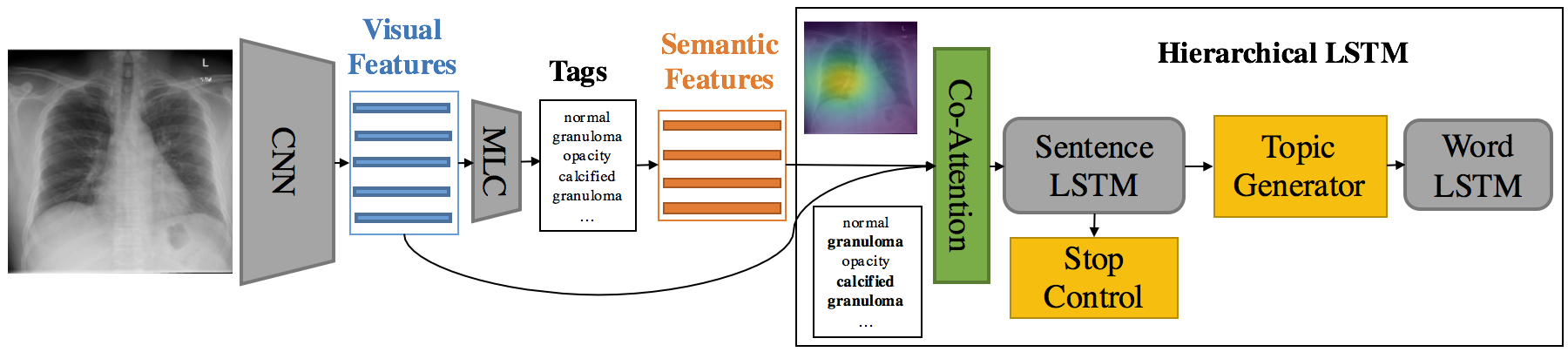}
  \caption{Illustration of the proposed model. MLC denotes a \ti{multi-label classification} network. Semantic features are the word embeddings of the predicted tags. The boldfaced tags ``calcified granuloma'' and ``granuloma'' are attended by the co-attention network.}
  \label{fig:model}
\end{figure*}

\section{Related Works}\label{related_work}

% ----- Automatic Medical Report Generation -----
\paragraph{Textual labeling of medical images}
\label{related_1}
There have been several works aiming at attaching ``texts" to medical images. In their settings, the target ``texts" are either fully-structured or semi-structured (e.g. tags, templates), rather than natural texts. \citet{kisilev2015medical} build a pipeline to predict the attributes of medical images. \citet{shin2016learning} adopt a CNN-RNN based framework to predict tags (e.g. locations, severities) of chest x-ray images. The work closest to ours is recently contributed by \cite{zhang2017mdnet}, which aims at generating semi-structured pathology reports, whose contents are restricted to 5 predefined topics. 

However, in the real-world, different physicians usually have different writing habits and different x-ray images will represent different abnormalities. Therefore, collecting semi-structured reports is less practical and thus it is important to build models to learn from natural reports. To the best of our knowledge, our work represents the first one that generates truly natural reports written by physicians, which are usually long and cover diverse topics.

%

% ----- Image Captioning with Deep Learning -----
\paragraph{Image captioning with deep learning}\label{related_2}
Image captioning aims at automatically generating text descriptions for given images. Most recent image captioning models are based on a CNN-RNN framework \cite{vinyals2015show,fang2015captions,karpathy2015deep,xu2015show,you2016image,Krause_2017_CVPR}.

Recently, attention mechanisms have been shown to be useful for image captioning \cite{xu2015show,you2016image}. \citet{xu2015show} introduce a spatial-visual attention mechanism over image features extracted from intermediate layers of the CNN. \citet{you2016image} propose a semantic attention mechanism over tags of given images. To better leverage both the visual features and semantic tags, we propose a co-attention mechanism for report generation.

Instead of only generating one-sentence caption for images, \citet{Krause_2017_CVPR} and \citet{Liang_2017_ICCV} generate paragraph captions using a hierarchical LSTM. Our method also adopts a hierarchical LSTM for paragraph generation, but unlike \citet{Krause_2017_CVPR}, we use a co-attention network to generate topics.

%!TEX root = ./acl_2018.tex
\pdfoutput=1
\section{Methods}\label{methods}

% ----- Overall Framework -----
\subsection{Overview}\label{overall_framework}
A complete diagnostic report for a medical image is comprised of both text descriptions (long paragraphs) and lists of tags, as shown in Figure~\ref{fig:example}. We propose a \ti{multi-task hierarchical model with co-attention} for automatically predicting keywords and generating long paragraphs. Given an image which is divided into regions, we use a CNN to learn visual features for these patches. Then these visual features are fed into a \ti{multi-label classification} (MLC) network to predict the relevant tags. In the tag vocabulary, each tag is represented by a word-embedding vector. Given the predicted tags for a specific image, their word-embedding vectors serve as the semantic features of this image. Then the visual features and semantic features are fed into a \ti{co-attention} model to generate a context vector that simultaneously captures the visual and semantic information of this image. As of now, the encoding process is completed. 

Next, starting from the context vector, the decoding process generates the text descriptions. The description of a medical image usually contains multiple sentences, and each sentence focuses on one specific topic. Our model leverages this compositional structure to generate reports in a hierarchical way: it first generates a sequence of high-level topic vectors representing sentences, then generates a sentence from each topic vector. Specifically, the context vector is inputted into a \ti{sentence LSTM}, which unrolls for a few steps and produces a topic vector at each step. A topic vector represents the semantics of a sentence to be generated. Given a topic vector, the \ti{word LSTM} takes it as input and generates a sequence of words to form a sentence. The termination of the unrolling process is controlled by the sentence LSTM.

% -----  
\subsection{Tag Prediction}

The first task of our model is predicting the tags of the given image. We treat the tag prediction task as a multi-label classification task. Specifically, given an image $I$, we first extract its features $\{\mathbf{v}_n\}_{n=1}^N\in\mathbb{R}^D$ from an intermediate layer of a CNN, and then feed $\{\mathbf{v}_n\}_{n=1}^N$ into a \emph{multi-label classification} (MLC) network to generate a distribution over all of the $L$ tags:

\begin{small}
\begin{equation}
\mathbf{p}_{\mathbf{l},pred}(\mathbf{l}_i=1|\{\mathbf{v}_n\}_{n=1}^N) \propto \exp(\text{MLC}_i(\{\mathbf{v}_n\}_{n=1}^N))
\end{equation}
\end{small}

\noindent where $\mathbf{l}\in\mathbb{R}^L$ is a tag vector, $\mathbf{l}_i=1 / 0$ denote the presence and absence of the $i$-th tag respectively, and MLC$_i$ means the $i$-th output of the MLC network.

For simplicity, we extract visual features from the last convolutional layer of the VGG-19 model \cite{Simonyan14c} and use the last two fully connected layers of VGG-19 for MLC.

Finally, the embeddings of the $M$ most likely tags $\{\mathbf{a}_m\}_{m=1}^M\in\mathbb{R}^E$ are used as semantic features for topic generation.

% ----- Co-Attention -----
\subsection{Co-Attention} Previous works have shown that visual attention alone can perform fairly well for localizing objects \cite{ba2014multiple} and aiding caption generation \cite{xu2015show}. However, visual attention does not provide sufficient high level semantic information. For example, only looking at the right lower region of the chest x-ray image (Figure \ref{fig:example}) without accounting for other areas, we might not be able to recognize what we are looking at, not to even mention detecting the abnormalities. In contrast, the tags can always provide the needed high level information. To this end, we propose a co-attention mechanism which can simultaneously attend to visual and semantic modalities.

In the sentence LSTM at time step $s$, the joint context vector $\mathbf{ctx}^{(s)}\in\mathbb{R}^C$ is generated by a co-attention network $f_{co_{att}}$($\{\mathbf{v}_n\}_{n=1}^N$, $\{\mathbf{a}_m\}_{m=1}^M$, $\mathbf{h}_{sent}^{(s-1)}$), where $\mathbf{h}_{sent}^{(s-1)}\in\mathbb{R}^H$ is the sentence LSTM hidden state at time step $s-1$. The co-attention network $f_{co_{att}}$ uses a single layer feedforward network to compute the soft visual attentions and soft semantic attentions over input image features and tags:

\begin{small}
\begin{align}
\alpha_{\mathbf{v},n} &\propto \exp(\mathbf{W}_{\mathbf{v}_{att}}\tanh(\mathbf{W}_\mathbf{v}\mathbf{v}_n + \mathbf{W}_{\mathbf{v},\mathbf{h}}\mathbf{h}_{sent}^{(s-1)}))\\
\alpha_{\mathbf{a},m} &\propto \exp(\mathbf{W}_{\mathbf{a}_{att}}\tanh(\mathbf{W}_\mathbf{a}\mathbf{a}_m + \mathbf{W}_{\mathbf{a},\mathbf{h}}\mathbf{h}_{sent}^{(s-1)}))
\end{align}
\end{small}

\noindent where $\mathbf{W}_\mathbf{v}$, $\mathbf{W}_{\mathbf{v},\mathbf{h}}$, and $\mathbf{W}_{\mathbf{v}_{att}}$ are parameter matrices of the visual attention network. $\mathbf{W}_\mathbf{a}$, $\mathbf{W}_{\mathbf{a},\mathbf{h}}$, and $\mathbf{W}_{\mathbf{a}_{att}}$ are parameter matrices of the semantic attention network. 

The visual and semantic context vectors are computed as:

\begin{small}
$$\mathbf{v}_{att}^{(s)} = \sum_{n=1}^N\alpha_{\mathbf{v},n}\mathbf{v}_n, \quad \mathbf{a}_{att}^{(s)} = \sum_{m=1}^M\alpha_{\mathbf{a},m}\mathbf{a}_m.$$
\end{small}

There are many ways to combine the visual and semantic context vectors such as concatenation and element-wise operations. In this paper, we first concatenate these two vectors as $[\mathbf{v}_{att}^{(s)}; \mathbf{a}_{att}^{(s)}]$, and then use a fully connected layer $\mathbf{W}_{fc}$ to obtain a joint context vector: 

\begin{small}
\begin{equation}
\mathbf{ctx}^{(s)} = \mathbf{W}_{fc}[\mathbf{v}_{att}^{(s)}; \mathbf{a}_{att}^{(s)}].
\end{equation}
\end{small}

% ----- Sentence LSTM -----
% \vspace{-0.5cm}
\subsection{Sentence LSTM}\label{sentence_lstm}
The sentence LSTM is a single-layer LSTM that takes the joint context vector $\mathbf{ctx}\in\mathbb{R}^C$ as its input, and generates topic vector $\mathbf{t}\in\mathbb{R}^K$ for word LSTM through topic generator and determines whether to continue or stop generating captions by a stop control component.

% - Topic Vector -
\paragraph{Topic generator}
We use a deep output layer \cite{pascanu2013construct} to strengthen the context information in topic vector $\mathbf{t}^{(s)}$, by combining the hidden state $\mathbf{h}_{sent}^{(s)}$ and the joint context vector $\mathbf{ctx}^{(s)}$ of the current step:
\begin{small}
\begin{equation}
\mathbf{t}^{(s)} = \tanh(\mathbf{W}_{\mathbf{t}, \mathbf{h}_{sent}}\mathbf{h}_{sent}^{(s)} + \mathbf{W}_{\mathbf{t}, \mathbf{ctx}}\mathbf{ctx}^{(s)})
\end{equation}
\end{small}

\noindent where $\mathbf{W}_{\mathbf{t}, \mathbf{h}_{sent}}$ and $\mathbf{W}_{\mathbf{t}, \mathbf{ctx}}$ are weight parameters.

% - Stop Control -
\paragraph{Stop control}
We also apply a deep output layer to control the continuation of the sentence LSTM. The layer takes the previous and current hidden state $\mathbf{h}_{sent}^{(s-1)}$, $\mathbf{h}_{sent}^{(s)}$ as input and produces a distribution over \{\emph{STOP}=1, \emph{CONTINUE}=0\}:

\begin{small}
\begin{equation}
\begin{split}
&p(STOP|\mathbf{h}_{sent}^{(s-1)}, \mathbf{h}_{sent}^{(s)})\propto\\
&\exp\{\mathbf{W}_{stop}\tanh(\mathbf{W}_{stop, s-1}\mathbf{h}_{sent}^{(s-1)}+\mathbf{W}_{stop, s}\mathbf{h}_{sent}^{(s)})\}
\end{split}
\end{equation}
\end{small}

\noindent where $\mathbf{W}_{stop}$, $\mathbf{W}_{stop, s-1}$ and $\mathbf{W}_{stop, s}$ are parameter matrices. If $p(STOP|\mathbf{h}_{sent}^{(s-1)}, \mathbf{h}_{sent}^{(s)})$ is greater than a predefined threshold (e.g. 0.5), then the sentence LSTM will stop producing new topic vectors and the word LSTM will also stop producing words.

% ----- Word LSTM -----
\subsection{Word LSTM} \label{word_lstm}
The words of each sentence are generated by a word LSTM. Similar to \cite{Krause_2017_CVPR}, the topic vector $\mathbf{t}$ produced by the sentence LSTM and the special \emph{START} token are used as the first and second input of the word LSTM, and the subsequent inputs are the word sequence.

The hidden state $\mathbf{h}_{word}\in\mathbb{R}^H$ of the word LSTM is directly used to predict the distribution over words:

\begin{small}
\begin{equation}
p(word|\mathbf{h}_{word})\propto \exp(\mathbf{W}_{out}\mathbf{h}_{word})
\end{equation}
\end{small}

\noindent where $\mathbf{W}_{out}$ is the parameter matrix. After each word-LSTM has generated its word sequences, the final report is simply the concatenation of all the generated sequences.

% ----- Model Learning -----
\subsection{Parameter Learning}
Each training example is a tuple ($I$, $\mathbf{l}$, $\mathbf{w}$) where $I$ is an image, $\mathbf{l}$ denotes the ground-truth tag vector, and $\mathbf{w}$ is the diagnostic paragraph, which is comprised of $S$ sentences and each sentence consists of $T_s$ words. 

Given a training example ($I$, $\mathbf{l}$, $\mathbf{w}$), our model first performs multi-label classification on $I$ and produces a distribution $\mathbf{p}_{\mathbf{l},pred}$ over all tags. Note that $\mathbf{l}$ is a binary vector which encodes the presence and absence of tags. We can obtain the ground-truth tag distribution by normalizing $\mathbf{l}$: $\mathbf{p_\mathbf{l}}=\mathbf{l}/||\mathbf{l}||_1$. The training loss of this step is a cross-entropy loss $\ell_{tag}$ between $\mathbf{p_\mathbf{l}}$ and $\mathbf{p}_{\mathbf{l},pred}$.

Next, the sentence LSTM is unrolled for $S$ steps to produce topic vectors and also distributions over \{\emph{STOP}, \emph{CONTINUE}\}: $p_{stop}^{s}$. Finally, the $S$ topic vectors are fed into the word LSTM to generate words $\mathbf{w}_{s,t}$. The training loss of caption generation is the combination of two cross-entropy losses: $\ell_{sent}$ over stop distributions $p_{stop}^{s}$ and $\ell_{word}$ over word distributions $p_{s,t}$. Combining the pieces together, we obtain the overall training loss:

\begin{small}
\begin{equation}
\begin{split}
\ell(I, \mathbf{l}, \mathbf{w}) &= \lambda_{tag}\ell_{tag} \\
&+ \lambda_{sent}\sum_{s=1}^S\ell_{sent}(p_{stop}^{s}, I\{s=S\}) \\
&+ \lambda_{word}\sum_{s=1}^{S}\sum_{t=1}^{T_s}\ell_{word}(p_{s,t}, w_{s,t})
\end{split}
\end{equation}
\end{small} 

In addition to the above training loss, there is also a regularization term for visual and semantic attentions. Similar to \cite{xu2015show}, let $\boldsymbol\alpha\in\mathbb{R}^{N\times S}$ and $\boldsymbol\beta\in\mathbb{R}^{M\times S}$ be the matrices of visual and semantic attentions respectively, then the regularization loss over $\boldsymbol\alpha$ and $\boldsymbol\beta$ is:

\begin{small}
\begin{equation}
\ell_{reg} = \lambda_{reg}[\sum_n^N(1-\sum_s^S \alpha_{n,s})^2 + \sum_m^M(1-\sum_s^S \beta_{m,s})^2]
\end{equation}
\end{small}

\noindent Such regularization encourages the model to pay equal attention over different image regions and different tags.

%!TEX root = ./acl_2018.tex
\pdfoutput=1
\section{Experiments}\label{experiments}
In this section, we evaluate the proposed model with extensive quantitative and qualitative experiments.

% ----- Datasets -----
\subsection{Datasets}

We used two publicly available medical image datasets to evaluate our proposed model. 
\paragraph{IU X-Ray} 
The Indiana University Chest X-Ray Collection (IU X-Ray) \cite{demner2015preparing} is a set of chest x-ray images paired with their corresponding diagnostic reports. The dataset contains 7,470 pairs of images and reports. Each report consists of the following sections: \emph{impression}, \emph{findings}, \emph{tags}\footnote{There are two types of tags: manually generated (MeSH) and Medical Text Indexer (MTI) generated.}, \emph{comparison}, and \emph{indication}. In this paper, we treat the contents in \emph{impression} and \emph{findings} as the target captions\footnote{The \emph{impression} and \emph{findings} sections are concatenated together as a long paragraph, since \emph{impression} can be viewed as a conclusion or topic sentence of the report.} to be generated and the Medical Text Indexer (MTI) annotated tags as the target tags to be predicted (Figure \ref{fig:example} provides an example). 

We preprocessed the data by converting all tokens to lowercases, removing all of non-alpha tokens, which resulting in 572 unique tags and 1915 unique words. On average, each image is associated with 2.2 tags, 5.7 sentences, and each sentence contains 6.5 words. Besides, we find that top 1,000 words cover 99.0\% word occurrences in the dataset, therefore we only included top 1,000 words in the dictionary. Finally, we randomly selected 500 images for validation and 500 images for testing.

\paragraph{PEIR Gross} 
The Pathology Education Informational Resource (PEIR) digital library\footnote{PEIR is \textcopyright University of Alabama at Birmingham, Department of Pathology. (http://peir.path.uab.edu/library/)} is a public medical image library for medical education. We collected the images together with their descriptions in the Gross sub-collection, resulting in the PEIR Gross dataset that contains 7,442 image-caption pairs from 21 different sub-categories. Different from the IU X-Ray dataset, each caption in PEIR Gross contains only one sentence. We used this dataset to evaluate our model's ability of generating single-sentence report.

For PEIR Gross, we applied the same preprocessing as IU X-Ray, which yields 4,452 unique words. On average, each image contains 12.0 words. Besides, for each caption, we selected 5 words with the highest tf-idf scores as tags.

% START-Table: main results
\begin{table*}[t!]
\centering
\scriptsize
\begin{tabular}{c|l|c c c c c c c}
\hline
Dataset & Methods & BLEU-1 & BLEU-2 & BLEU-3 & BLEU-4 & METEOR & ROUGE & CIDER\\
\hline
\multirow{8}{*}{IU X-Ray}
 & CNN-RNN \cite{vinyals2015show}   & 0.316 & 0.211 & 0.140 & 0.095 & 0.159 & 0.267 & 0.111\\
 & LRCN \cite{donahue2015long}       & 0.369 & 0.229 & 0.149 & 0.099 & 0.155 & 0.278 & 0.190\\
 & Soft ATT \cite{xu2015show}        & 0.399 & 0.251 & 0.168 & 0.118 & 0.167 & 0.323 & 0.302\\
 & ATT-RK \cite{you2016image}        & 0.369 & 0.226 & 0.151 & 0.108 & 0.171 & 0.323 & 0.155\\
 \cline{2-9}
 & Ours-no-Attention   & 0.505 & 0.383 & 0.290 & 0.224 & 0.200 & 0.420 & 0.259\\
 & Ours-Semantic-only  & 0.504 & 0.371 & 0.291 & 0.230 & 0.207 & 0.418 & 0.286\\
 & Ours-Visual-only    & 0.507 & 0.373 & 0.297 & 0.238 & 0.211 & 0.426 & 0.300\\
 & Ours-CoAttention    & \textbf{0.517} & \textbf{0.386} & \textbf{0.306} & \textbf{0.247} & \textbf{0.217} & \textbf{0.447} & \textbf{0.327}\\
\hline
\hline
\multirow{7}{*}{PEIR Gross}
 & CNN-RNN \cite{vinyals2015show}   & 0.247 & 0.178 & 0.134 & 0.092 & 0.129 & 0.247 & 0.205\\
 & LRCN \cite{donahue2015long}      & 0.261 & 0.184 & 0.136 & 0.088 & 0.135 & 0.254 & 0.203\\
 & Soft ATT \cite{xu2015show}       & 0.283 & 0.212 & 0.163 & 0.113 & 0.147 & 0.271 & 0.276\\
 & ATT-RK  \cite{you2016image}       & 0.274 & 0.201 & 0.154 & 0.104 & 0.141 & 0.264 & 0.279\\
 \cline{2-9}
 & Ours-No-Attention   & 0.248 & 0.180 & 0.133 & 0.093 & 0.131 & 0.242 & 0.206\\
 & Ours-Semantic-only  & 0.263 & 0.191 & 0.145 & 0.098 & 0.138 & 0.261 & 0.274\\
 & Ours-Visual-only    & 0.284 & 0.209 & 0.156 & 0.105 & \textbf{0.149} & 0.274 & 0.280\\
 & Ours-CoAttention    & \textbf{0.300} & \textbf{0.218} & \textbf{0.165} & \textbf{0.113} & \textbf{0.149} & \textbf{0.279} & \textbf{0.329}\\
\hline
\end{tabular}
\caption{Main results for paragraph generation on the IU X-Ray dataset (upper part), and single sentence generation on the PEIR Gross dataset (lower part). BLUE-n denotes the BLEU score that uses up to n-grams.}
\label{tab:main_paragraph}
% \vspace{0.8cm}
\end{table*}
% END-Table: main results

% ----- Implementation Details -----
\subsection{Implementation Details}
We used the full VGG-19 model \cite{Simonyan14c} for tag prediction. As for the training loss of the multi-label classification (MLC) task, since the number of tags for semantic attention is fixed as 10, we treat MLC as a multi-label retrieval task and adopt a softmax cross-entropy loss (a multi-label ranking loss), similar to~\cite{gong2013deep,guillaumin2009tagprop}.  

In paragraph generation, we set the dimensions of all hidden states and word embeddings as 512. For words and tags, different embedding matrices were used since a tag might contain multiple words. We utilized the embeddings of the 10 most likely tags as the semantic feature vectors $\{\mathbf{a}_m\}_{m=1}^{M=10}$. We extracted the visual features from the last convolutional layer of the VGG-19 network, which yields a $14\times14\times512$ feature map. 

We used the Adam~\cite{kingma2014adam} optimizer for parameter learning. The learning rates for the CNN (VGG-19) and the hierarchical LSTM were 1e-5 and 5e-4 respectively. The weights ($\lambda_{tag}$, $\lambda_{sent}$, $\lambda_{word}$ and $\lambda_{reg}$) of different losses were set to 1.0. The threshold for stop control was 0.5. Early stopping was used to prevent over-fitting. 

% ----- Baselines -----
\subsection{Baselines}
We compared our method with several state-of-the-art image captioning models: CNN-RNN \cite{vinyals2015show}, LRCN \cite{donahue2015long}, Soft ATT \cite{xu2015show}, and ATT-RK \cite{you2016image}. We re-implemented all of these models and adopt VGG-19 \cite{Simonyan14c} as the CNN encoder. Considering these models are built for single sentence captions and to better show the effectiveness of the hierarchical LSTM and the attention mechanism for paragraph generation, we also implemented a hierarchical model without any attention: Ours-no-Attention. The input of Ours-no-Attention is the overall image feature of VGG-19, which has a dimension of 4096. Ours-no-Attention can be viewed as a CNN-RNN \cite{vinyals2015show} equipped with a hierarchical LSTM decoder. To further show the effectiveness of the proposed co-attention mechanism, we also implemented two ablated versions of our model: Ours-Semantic-only and Ours-Visual-only, which takes solely the semantic attention or visual attention context vector to produce topic vectors.

% START- Figure: paragraph generation
\begin{figure*}[h!]
  \centering
  \scriptsize
  \includegraphics[width=\textwidth]{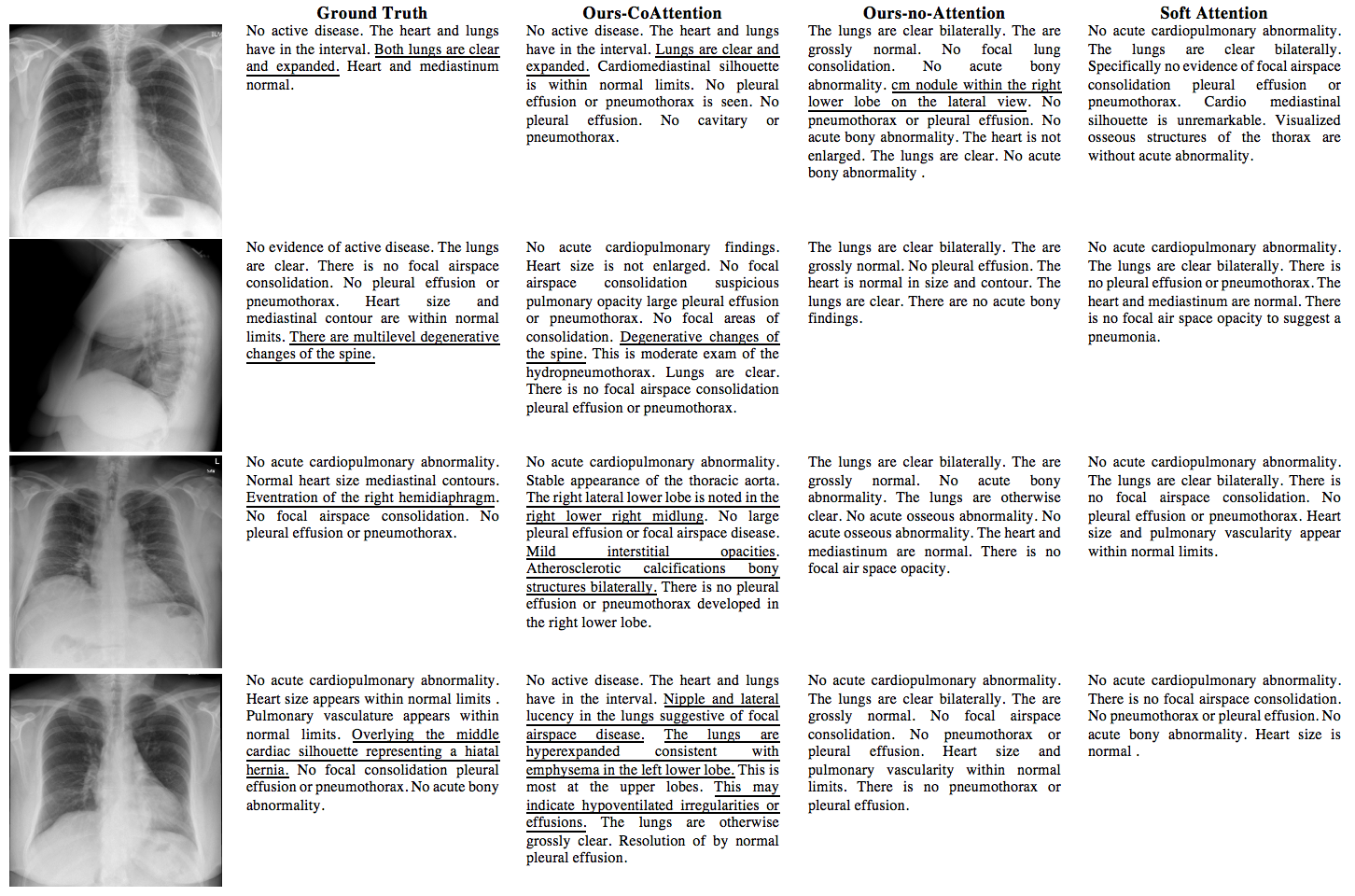}
  \caption{Illustration of paragraph generated by Ours-CoAttention, Ours-no-Attention, and Soft Attention models. The underlined sentences are the descriptions of detected abnormalities. The second image is a lateral x-ray image. Top two images are positive results, the third one is a partial failure case and the bottom one is failure case. These images are from test dataset.}
  \label{fig:paragraph_generation}
\end{figure*}
% END- Figure: paragraph generation

% ----- Main Results -----
\subsection{Quantitative Results}
We report the paragraph generation (upper part of Table \ref{tab:main_paragraph}) and one sentence generation (lower part of Table \ref{tab:main_paragraph}) results using the standard image captioning evaluation tool \footnote{https://github.com/tylin/coco-caption} which provides evaluation on the following metrics: BLEU \cite{papineni2002bleu}, METEOR \cite{denkowski2014meteor}, ROUGE \cite{lin2004rouge}, and CIDER \cite{vedantam2015cider}. 

For paragraph generation, as shown in the upper part of Table \ref{tab:main_paragraph}, it is clear that models with a single LSTM decoder perform much worse than those with a hierarchical LSTM decoder. Note that the only difference between Ours-no-Attention and CNN-RNN \cite{vinyals2015show} is that Ours-no-Attention adopts a hierarchical LSTM decoder while CNN-RNN \cite{vinyals2015show} adopts a single-layer LSTM. The comparison between these two models directly demonstrates the effectiveness of the hierarchical LSTM. This result is not surprising since it is well-known that a single-layer LSTM cannot effectively model long sequences \cite{liu2015multi,martin2017parallelizing}. Additionally, employing semantic attention alone (Ours-Semantic-only) or visual attention alone (Ours-Visual-only) to generate topic vectors does not seem to help caption generation a lot. The potential reason might be that visual attention can only capture the visual information of sub-regions of the image and is unable to correctly capture the semantics of the entire image. Semantic attention is inadequate of localizing small abnormal image-regions. Finally, our full model (Ours-CoAttention) achieves the best results on all of the evaluation metrics, which demonstrates the effectiveness of the proposed co-attention mechanism.

For the single-sentence generation results (shown in the lower part of Table \ref{tab:main_paragraph}), the ablated versions of our model (Ours-Semantic-only and Ours-Visual-only) achieve competitive scores compared with the state-of-the-art methods. Our full model (Ours-CoAttention) outperforms all of the baseline, which indicates the effectiveness of the proposed co-attention mechanism. 

% - Co-att image
\begin{figure*}[ht!]
  \centering
  \scriptsize
  \includegraphics[width=\textwidth]{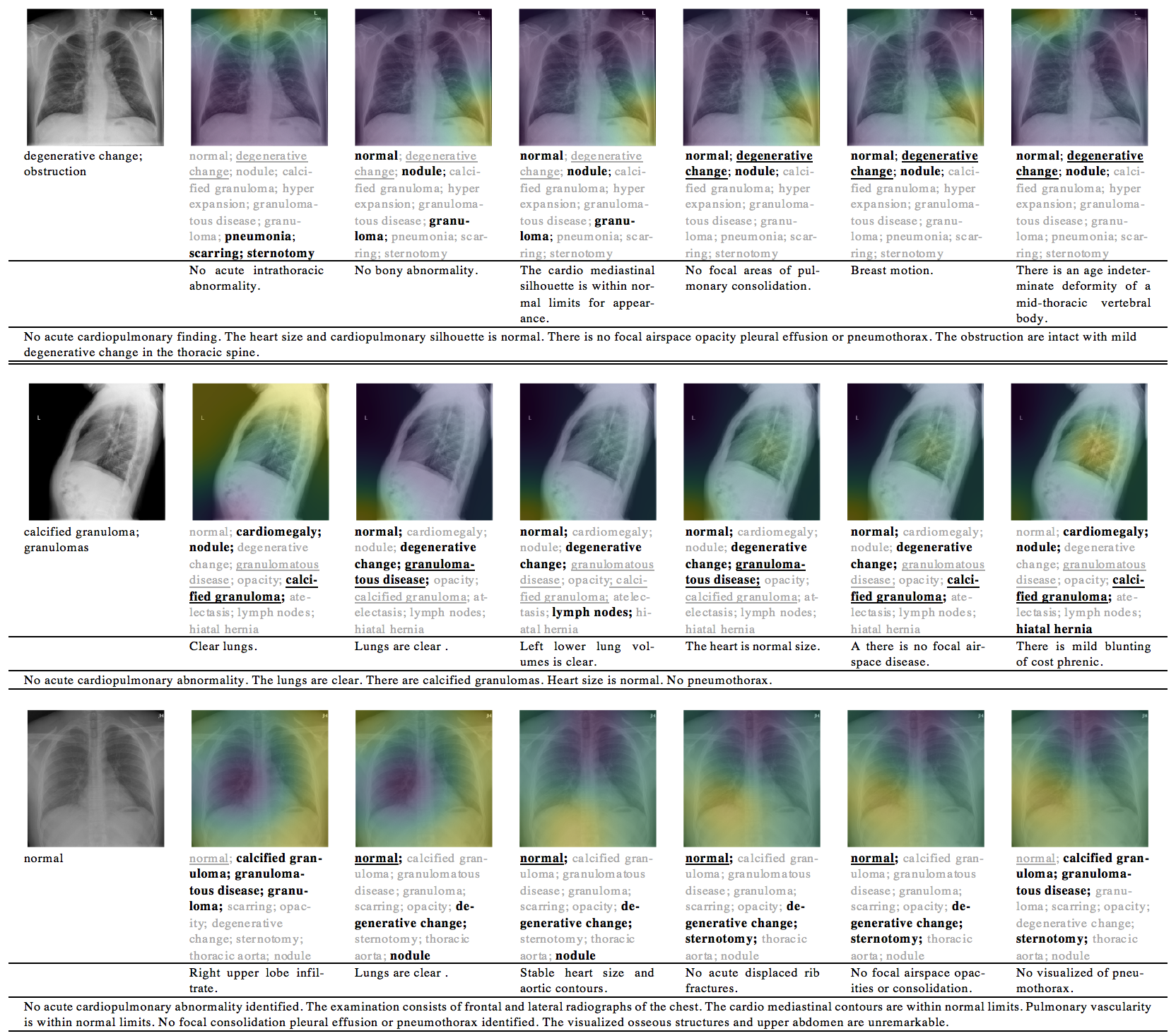}
  \caption{Visualization of co-attention for three examples. Each example is comprised of four things: (1) image and visual attentions; (2) ground truth tags and semantic attention on predicted tags; (3) generated descriptions; (4) ground truth descriptions. For the semantic attention, three tags with highest attention scores are highlighted. The underlined tags are those appearing in the ground truth.}
  \label{fig:co_att}
  % \vspace{-0.35cm}
\end{figure*}

% ----- Qualitative Results -----
\subsection{Qualitative Results}
% - Paragraph Generation -
\subsubsection{Paragraph Generation}
An illustration of paragraph generation by three models (Ours-CoAttention, Ours-no-Attention and Soft Attention models) is shown in Figure \ref{fig:paragraph_generation}. We can find that different sentences have different topics. The first sentence is usually a high level description of the image, while each of the following sentences is associated with one area of the image (e.g. ``lung'', ``heart''). Soft Attention and Ours-no-Attention models detect only a few abnormalities of the images and the detected abnormalities are incorrect. In contrast, Ours-CoAttention model is able to correctly describe many true abnormalities (as shown in top three images). This comparison demonstrates that  co-attention is better at capturing abnormalities.

For the third image, Ours-CoAttention model successfully detects the area (``right lower lobe'') which is abnormal (``eventration''), however, it fails to precisely describe this abnormality. In addition, the model also finds abnormalities about ``interstitial opacities'' and ``atheroscalerotic calcification'', which are not considered as true abnormality by human experts. The potential reason for this mis-description might be that this x-ray image is darker (compared with the above images), and our model might be very sensitive to this change. 

The image at the bottom is a failure case of Ours-CoAttention. However, even though the model makes the wrong judgment about the major abnormalities in the image, it does find some unusual regions: ``lateral lucency'' and ``left lower lobe''. 

To further understand models' ability of detecting abnormalities, we present the portion of sentences which describe the normalities and abnormalities in Table \ref{tab:protion_abn}. We consider sentences which contain ``no'', ``normal'', ``clear'', ``stable'' as sentences describing normalities. It is clear that Ours-CoAttention best approximates the ground truth distribution over normality and abnormality. 

\begin{table}[t]
\centering
\scriptsize
\begin{tabular}{c|c c c}
\hline
Method & Normality & Abnormality & Total\\
\hline
Soft Attention     & 0.510 & 0.490 & 1.0\\
Ours-no-Attention  & 0.753 & 0.247 & 1.0\\
Ours-CoAttention   & 0.471 & 0.529 & 1.0\\
\hline
Ground Truth       & 0.385 & 0.615 & 1.0\\
\hline
\end{tabular}
\caption{Portion of sentences which describe the normalities and abnormalities in the image.} \label{tab:protion_abn}
% \vspace{-0.4cm}
\end{table}

% - Co-Attention -
\subsubsection{Co-Attention Learning} \label{co_att}
Figure \ref{fig:co_att} presents visualizations of co-attention. The first property shown by Figure \ref{fig:co_att} is that the sentence LSTM can generate different topics at different time steps since the model focuses on different image regions and tags for different sentences. The next finding is that visual attention can guide our model to concentrate on relevant regions of the image. For example, the third sentence of the first example is about ``cardio'', and the visual attention concentrates on regions near the heart. Similar behavior can also be found for semantic attention: for the last sentence in the first example, our model correctly concentrates on ``degenerative change'' which is the topic of the sentence. Finally, the first sentence of the last example presents a mis-description caused by incorrect semantic attention over tags. Such incorrect attention can be reduced by building a better tag prediction module.

\section{Conclusion}\label{conclusion}

In this paper, we study how to automatically generate textual reports for medical images, with the goal to help medical professionals produce reports more accurately and efficiently. Our proposed methods address three major challenges: (1) how to generate multiple heterogeneous forms of information within a unified framework, (2) how to localize abnormal regions and produce accurate descriptions for them, (3) how to generate long texts that contain multiple sentences or even paragraphs. To cope with these challenges, we propose a multi-task learning framework which jointly predicts tags and generates descriptions. We introduce a co-attention mechanism that can simultaneously explore visual and semantic information to accurately localize and describe abnormal regions. We develop a hierarchical LSTM network that can more effectively capture long-range semantics and produce high quality long texts. On two medical datasets containing radiology and pathology images, we demonstrate the effectiveness of the proposed methods through quantitative and qualitative studies. 

%\bibliography{acl2018}
\bibliography{acl2018r}
\bibliographystyle{acl_natbib}

\end{document}